\documentclass{iopconfser}
\usepackage{amsmath}
\usepackage{amssymb}
\usepackage{graphicx}
\usepackage[normalem]{ulem}
\usepackage{url}
\begin{document}

\title{Reinforcement Learning for Synchronised Flow Control in a Dual-Gate Resin Infusion System}

\setlength{\parskip}{1em}

\author{Miguel Camacho-S\'anchez$^{1}$, Fernando Garc\'ia-Torres$^{1}$, Jesper John Lisegaard$^{2}$, Roc\'io del Amor$^{1,3}$, Sankhya Mohanty$^{2}$ and Valery Naranjo$^{1,3}$}

\affil{$^1$Instituto Universitario de Investigaci\'on en Tecnolog\'ia Centrada en el Ser Humano, HUMAN-tech, Universitat Polit\`ecnica de Val\`encia, Valencia, Spain.}
\affil{$^2$ Department of Civil and Mechanical Engineering, Technical University of Denmark, Lyngby, Denmark.}
\affil{$^3$Artikode Intelligence S.L, Valencia, Spain.}

\email{micasan4@teleco.upv.es, vnaranjo@dcom.upv.es}

\begin{abstract}
Resin infusion (RI) and resin transfer moulding (RTM) are  critical processes for the manufacturing of high-performance fibre-reinforced polymer composites, particularly for large-scale applications such as wind turbine blades. Controlling the resin flow dynamics in these processes is critical to ensure the uniform impregnation of the fibre reinforcements, thereby preventing residual porosities and dry spots that impact the consequent structural integrity of the final component. This paper presents a reinforcement learning (RL) based strategy, established using process simulations, for synchronising the different resin flow fronts in an infusion scenario involving two resin inlets and a single outlet. Using Proximal Policy Optimisation (PPO), our approach addresses the challenge of managing the fluid dynamics in a partially observable environment. The results demonstrate the effectiveness of the RL approach in achieving an accurate flow convergence, highlighting its potential towards improving process control and product quality in composites manufacturing.
\end{abstract}

\section{Introduction}
Resin Infusion (RI) and Resin Transfer Moulding (RTM) are closed-mould manufacturing techniques widely adopted for producing high-performance fibre-reinforced polymer composites \cite{bickerton_advanced_2013}. They are especially suited for large structural components, such as wind turbine blades (WTBs), where weight reduction and mechanical strength are critical \cite{huberty_state_2024, hindersmann_confusion_2019}. RI processes offer significant advantages over other open-mould techniques, including higher and more consistent fibre volume fractions, improved structural properties, and scalability to large parts \cite{sainz_new_2015, branner_towards_2021}. However, manufacturing large and complex components introduces substantial challenges related to controlling the resin flow during the filling and curing stages \cite{yenilmez_variation_2009}.

A central difficulty in RI/RTM lies in ensuring the uniform resin impregnation of the dry fibre preform to prevent voids, dry spots, and thickness inconsistencies. These issues often arise from the pressure imbalances across inlets, the local variations in preform compaction and the resin viscosity \cite{kazmi_control_2019, govignon_experimental_2013}. Although process variants have been developed to improve control and reduce variability \cite{van_oosterom_objective_2019}, the lack of automated feedback mechanisms currently results in operators still relying on experience and trial-and-error to tune the process. As a result, simulation-based tools and intelligent control strategies are gaining traction in the field \cite{struzziero_multi-objective_2019}. Recent advances in flow monitoring using embedded sensor networks \cite{liu_monitoring_2021, dimassi_using_2021} motivate the integration of real-time, observation-driven control strategies to improve the quality and repeatability of RI/RTM processes. 

Reinforcement Learning (RL) has been reviewed extensively in broader manufacturing and production-control contexts \cite{nian_review_2020}. Cadavid et al. \cite{usuga_cadavid_machine_2020} presented a systematic overview of machine learning (ML) driven production planning and control, noting that RL is emerging for adaptive scheduling, logistics, and machine-level control, often integrated with sensor-rich environments. Additionally, Bertolini et al. \cite{bertolini_machine_2021} studied industrial ML applications, underscoring the value of RL for real-time decision-making in data-driven manufacturing systems. Panzer et al. surveyed different RL approaches applied to industrial processes such as the maintenance of facilities, energy management or production scheduling, among others. 

On the development and implementation of RL for RI and RTM process control,  RL has been applied recently to complex process control and flow systems like RTM and liquid composite moulding. Szarski et al. applied deep RL to optimise flow media placement in resin infusion, significantly reducing fill time by 32\% on aerospace composites when compared to expert-designed layouts—demonstrating RL potential in resin flow tasks \cite{szarski_instant_2023}. Previous studies have leveraged optimisation algorithms to determine the optimal placement of inlets and vents, aiming to enhance flow uniformity and reduce infusion time in RTM scenarios \cite{mathur_use_1999}. Complementary to these approaches, other works have explored advanced control techniques in resin infusion from a thermal management perspective. For example, Zhang et al \cite{zhang_fast_2024} presents a deep convolutional and recurrent neural network model to predict the spatio-temporal temperature distribution during the vacuum-assisted resin infusion moulding process, enabling rapid and accurate thermal monitoring for improved process control.

This work addresses a simplified yet representative case of the RI process: a horizontal rectangular mould with two opposing resin inlets—one at the top and one at the bottom. When resin is injected simultaneously from both ends, the two flow fronts advance toward each other. The control objective is to synchronise the injection rates such that the flow fronts meet precisely at the midpoint of the domain, where the vent of the system is located. Achieving this is considered key to minimising the internal defects and ensuring a uniform laminate quality.
In this work, a simulated system is configured with a limited set of virtual sensors placed at fixed spatial locations, providing observational data related to the flow front positions. Using an RL algorithm, the objective is to deduce the state of the system from the sensor array information and accordingly adjust the flow rates in real time. 

This setup poses several challenges: the non-linearity of the fluid dynamics, the delayed control effects and the need for close temporal coordination between spatially distant inlets. To overcome these obstacles, we use Proximal Policy Optimisation (PPO) - a state-of-the-art policy gradient algorithm known for stability and sample efficiency in continuous action domains - to learn a robust control policy under partial observability. The problem is formulated as a partially observable Markov decision process (POMDP), where the agent selects injection rates based on limited sensor feedback.

To explore the feasibility of RL-based control in the proposed environment, we investigate the performance of a PPO agent trained in this simulated two-input vertical configuration. The study focuses on the agent's ability to infer system state from sparse sensor feedback and to coordinate injection rates in real time.   To the best of our knowledge, there is no study of RL algorithms for the control and synchronisation of two flow fronts opposing each other in RTM processes.

\section{Problem Formulation}
\subsection{System Description}

The layout corresponds to the infusion system described in~\cite{Lisegaard2025}, consisting of a rectangular domain divided into an upper and lower region, each equipped with a dedicated inlet gate. These gates allow the injection of the resin into the system from both the top and bottom edges, respectively. 

In the investigated scenario, the lower region presents a higher flow rate compared to the upper region, primarily due to the differences in the local permeability. This intrinsic asymmetry introduces a nontrivial control challenge, requiring the RL agent to adapt its policy to counteract the natural imbalance. In addition to this heterogeneous permeability, the system contains artificial high-permeability channels, representing the geometrical features of the underlying non-fibre material (e.g. the cuts and grooves of the balsa wood layup), that lead to the well-known racetracking phenomena (for the remainder of this paper, we refer to these as \textit{racetracks}). These linear structures enable significantly faster flow along predefined paths and play a critical role in influencing the overall dynamics of the resin front. The presence of racetracks introduces anisotropic flow behaviour, further increasing the complexity of the control task.

The investigated domain includes a total of 90 virtual sensors uniformly distributed across the area, arranged in a grid of 6 columns and 15 rows. These sensors provide sparse, real-time information about the progression of the resin front.

\subsection{Objective}
The objective of the RL agent is to synchronise the arrival of the two resin fronts, starting from the top and bottom gates, such that they reach the central convergence region at approximately the same time. A successful policy should compensate for the inherent asymmetry in the environment (e.g., higher permeability in the lower region) and learn a temporal coordination strategy to ensure the simultaneous arrival of both flows.

\begin{figure}[ht]
    \centering
    \includegraphics[width=0.7\linewidth]{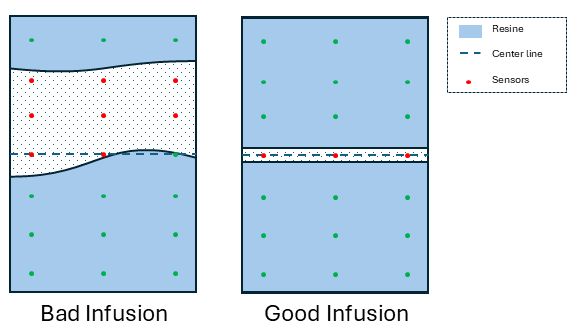}
    \caption{Comparison between two flow control outcomes. Left: unsynchronised arrival resulting in imbalance and delayed filling. Right: synchronised resin fronts converging at the center. }
    \label{fig:sync_comparison}
\end{figure}

Figure~\ref{fig:sync_comparison} illustrates the objective of the RL algorithm. The left panel shows a poorly synchronised case, where one flow significantly outpaces the other, resulting in delayed convergence and suboptimal filling dynamics. The right panel shows an example of a well-synchronised episode, where both fronts reach the central region nearly simultaneously, avoiding the formation of a dry spot. 
Throughout training, the agent is implicitly encouraged to align the timing of the two fronts via the reward signal, whose specific formulation is described in Section~\ref{sec:reward}.

\section{Methodology}

\subsection{Environment and Agent Setup}

RL problems are typically modelled as Markov Decision Processes (MDPs), defined by the tuple $(\mathcal{S}, \mathcal{A}, P, \- r, \gamma)$, where:

\begin{itemize}
    \item $\mathcal{S}$ is the state space, representing all possible configurations of the environment.
    \item $\mathcal{A}$ is the action space, containing all possible actions the agent can take.
    \item $P: \mathcal{S} \times \mathcal{A} \times \mathcal{S} \rightarrow [0, 1]$ is the transition probability function, with $P(s_{t+1} | s_t, a_t)$ denoting the probability of transitioning to state $s_{t+1}$ after taking action $a_t$ in state $s_t$.
    \item $r: \mathcal{S} \times \mathcal{A} \rightarrow \mathbb{R}$ is the reward function, which provides feedback to the agent after each interaction with the environment.
    \item $\gamma \in [0, 1)$ is the discount factor, which determines the importance of future rewards.
\end{itemize}

The simulation environment is based on the infusion system described in~\cite{Lisegaard2025} and consists of a two-dimensional rectangular domain divided into upper and lower regions, each with a dedicated inlet gate. At each timestep, resin is infused into the domain through these gates and propagates through a heterogeneous permeability field toward a central convergence region.

To encourage policy generalisation, a new environment is sampled at the beginning of every episode. The permeability field is randomly generated using a spatially correlated process, and the lower half of the domain is always assigned a higher average permeability than the upper half. This intrinsic asymmetry creates a consistent imbalance in flow velocity that the agent must learn to compensate for. Additionally, three high-permeability racetracks are inserted into the domain at random positions, orientations, and lengths. These structures introduce anisotropic flow patterns, increasing the complexity of the task and the diversity of training scenarios.

The agent perceives the environment through 90 virtual binary sensors uniformly distributed in a $6 \times 15$ grid. Each sensor outputs a value of 1 if it has been reached by the resin, and 0 otherwise. These values form a binary vector $o_t \in \mathcal{S}$ that provides a partial observation of the environment. Based on this input, the agent selects an action $a_t \in \mathcal{A}$ from a discrete action space consisting of three options: opening only the top gate, opening only the bottom gate, or opening both gates simultaneously. The option to close both gates is deliberately excluded to ensure that the resin continues to flow and to avoid stagnation.

Each episode terminates when all sensors located in a narrow horizontal band at the center of the domain are activated by the resin. This condition ensures that the task is evaluated only once both fronts have had the opportunity to reach the target region.

\begin{figure}[h!]
    \centering
    \includegraphics[width=\linewidth]{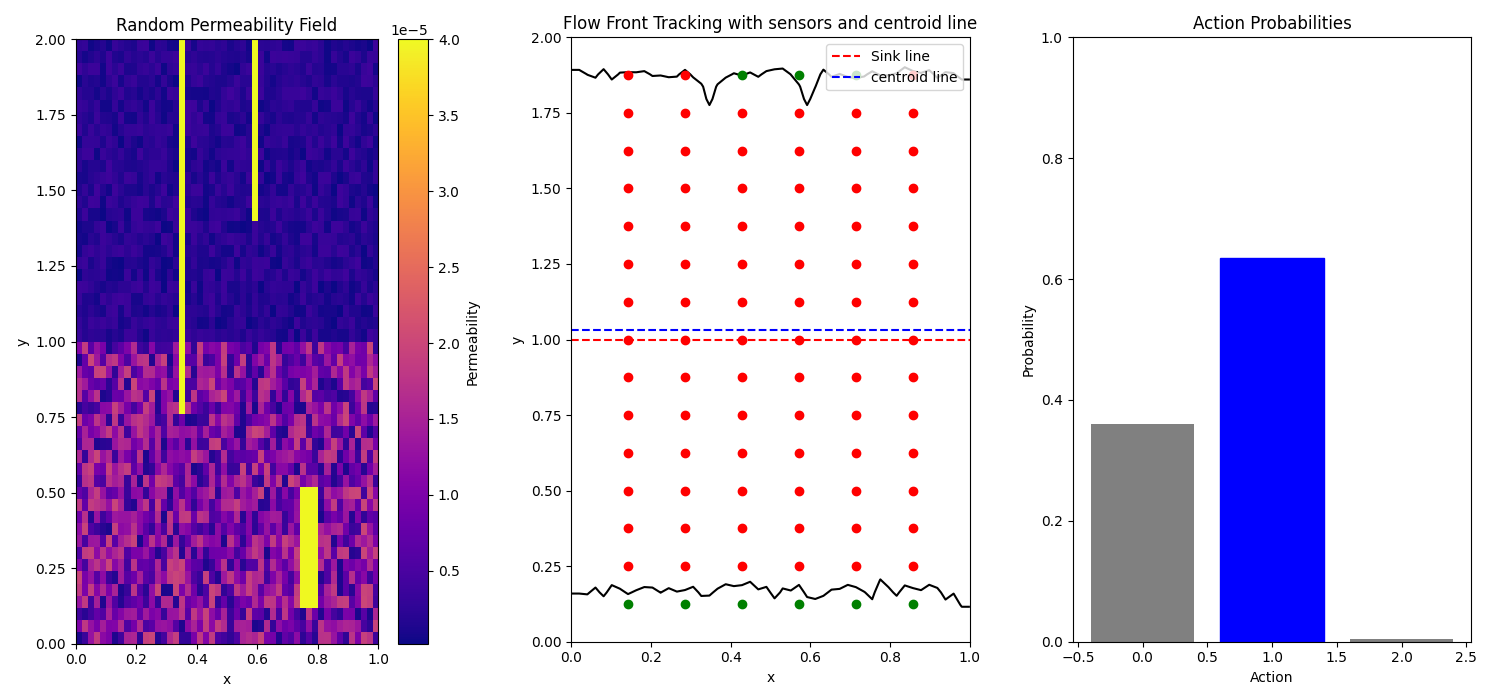}
    \caption{
    Overview of the simulation environment. From left to right: permeability field with randomly generated racetracks, pressure distribution, flow front and activated sensors with centroid line, and action probability distribution produced by the agent's policy.
    }
    \label{fig:env_overview}
\end{figure}

Figure~\ref{fig:env_overview} provides a visual overview of the environment. From left to right, it shows the sampled permeability field with racetracks, the pressure field, the flow front with activated sensors and centroid line, and the agent’s action probability distribution at a given timestep.

\subsection{Reinforcement Learning Algorithm}

The agent's objective is to learn a policy $\pi_\theta(a | s)$, parameterized by $\theta$, that maximizes the expected discounted return:
\begin{equation}
    J(\theta) = \mathbb{E}_{\pi_\theta} \left[ \sum_{t=0}^{\infty} \gamma^t r(s_t, a_t) \right].
\end{equation}

To optimize this objective, we use Proximal Policy Optimization ~\cite{schulman2017proximalpolicyoptimizationalgorithms}, a state-of-the-art policy gradient method that balances sample efficiency and stability. PPO improves upon standard policy gradient approaches by introducing a surrogate objective that penalises large deviations from the current policy.

Overall, PPO strikes a balance between exploration and exploitation by allowing multiple policy updates per batch of data while constraining the policy change through the clipped objective. This approach enables the agent to explore new actions without deviating too dramatically from learned behavior, preserving training stability. Compared to other policy gradient methods, PPO is simpler to implement, avoids second-order computations, and has demonstrated strong empirical performance and sample efficiency across both continuous and discrete control benchmarks~\cite{engstrom_implementation_2019}.

The clipped surrogate objective is defined as:
\begin{equation}
    L^{\text{CLIP}}(\theta) = \mathbb{E}_t \left[ 
        \min \left( r_t(\theta) \hat{A}_t, 
        \; \text{clip}(r_t(\theta), 1 - \epsilon, 1 + \epsilon) \hat{A}_t \right)
    \right],
\end{equation}
where:
\begin{itemize}
    \item $r_t(\theta) = \frac{\pi_\theta(a_t | s_t)}{\pi_{\theta_{\text{old}}}(a_t | s_t)}$ is the probability ratio between the new and old policies.
    \item $\hat{A}_t$ is an estimator of the advantage function at timestep $t$, which quantifies the relative value of action $a_t$ in state $s_t$.
    \item $\epsilon$ is a hyperparameter that controls the maximum allowable policy update, typically set to 0.1 or 0.2.
\end{itemize}

The use of the $\min$ operator and the clipping function ensures that policy updates remain within a bounded region, preventing excessively large steps that could destabilize training. This allows for multiple epochs of policy optimization using the same batch of data.

The advantage estimator $\hat{A}_t$ is commonly computed using Generalized Advantage Estimation (GAE) \cite{schulman2018highdimensionalcontinuouscontrolusing}:

\begin{equation}
    \hat{A}_t = \sum_{l=0}^{T-t} (\gamma \lambda)^l \delta_{t+l},
\end{equation}
where the temporal-difference residual is given by:
\begin{equation}
    \delta_t = r_t + \gamma V(s_{t+1}) - V(s_t),
\end{equation}
and $\lambda \in [0, 1]$ is a hyperparameter that trades off bias and variance in the estimator.

\subsection{Reward Function}
\label{sec:reward}
The design of the reward function plays a crucial role in guiding the learning process toward synchronized flow control. In our setup, the agent receives rewards only when a new symmetric row of sensors is activated, i.e., a pair of sensor rows equidistant from the centerline of the domain become fully activated during the same action step. This reward structure implicitly encourages the agent to activate the domain in a balanced manner from both sides.

To aid in the design of the reward function, we use the centroid of the activated sensor region as a geometric metric to characterize flow symmetry. It is computed from the spatial coordinates of all sensors that have been activated by the resin and represents the average position of the flow front. In particular, the vertical component of the centroid provides a useful indicator of imbalance between the top and bottom flows.

We experimented with two distinct reward functions:

\paragraph*{1) Symmetric Row Activation Reward}

In the first formulation, the agent receives a positive reward of $+1$ for each new symmetric row that becomes fully activated at a given timestep. If only one half of a row is activated (e.g., due to opening only the top gate), no reward is given. This reward encourages balanced progression from both gates but does not explicitly penalize asymmetry.

\paragraph*{2) Symmetric Activation + Centroid Penalty}

The second reward builds on the first by adding a penalty term that reflects how far the centroid deviates from the vertical centerline. At the end of each episode, a negative reward proportional to the final vertical distance of the centroid from the domain center is applied. Formally, if $c_y$ is the vertical coordinate of the centroid and $c_{\text{center}}$ is the centerline position, then the penalty is given by:

\begin{equation}
    r_{\text{penalty}} = -\alpha \cdot |c_y - c_{\text{center}}|,
\end{equation}

where $\alpha > 0$ is a scaling factor. This formulation encourages not only symmetric activation at each timestep, but also overall flow alignment by discouraging persistent asymmetries.

Both reward functions are normalised such that the final episode score lies within $[0, 1]$, facilitating fair comparison of learning performance across training runs.
\subsection{Performance Metrics}
The primary performance metric used to evaluate the agent is the cumulative reward obtained during an episode, henceforth referred to as the episode score. While several reward function designs are explored, all variants are normalised such that the episode score lies within the interval $[0, 1]$.

A score close to $+1$ indicates near-perfect synchronisation of the two resin fronts, implying successful coordination and task completion. On the contrary, scores near $0$ reflect poor synchronisation and failure to achieve the objective. This normalisation enables direct comparison of learning progress across different training configurations and reward structures.

Performance is tracked over time by monitoring the average episode score across multiple evaluation episodes. Once the agent consistently achieves scores near the upper bound, we consider it to have effectively solved the environment.
\subsection{Training Details}

The agent was trained using the PPO algorithm with a fixed set of hyperparameters across all experiments. The discount factor was set to $\gamma = 0.999$, encouraging long-term optimisation. GAE was used with a decay factor of $\lambda = 0.95$ to control the bias-variance tradeoff in advantage computation. A clipping threshold of $\epsilon = 0.1$ was used in the surrogate objective to ensure stable policy updates.

Each training batch was composed of $N = 32$ complete episodes, with each episode consisting of approximately 60–80 timesteps depending on the dynamics of the flow and the environment configuration. The policy was updated for $n_{\text{epochs}} = 4$ epochs per batch. Training was run for a total of $n_{\text{games}} = 2000$ episodes, which required approximately three hours of wall-clock time.

The simulation environment was executed on a single CPU, while policy updates were performed on a GPU after each batch of $N$ episodes. This hybrid setup allowed for efficient simulation sampling while leveraging GPU acceleration for neural network optimization. All hyperparameters were kept fixed across training runs, and no reward scaling or curriculum learning was applied.

\section{Results and Discussion}
\subsection{Quantitative Results}

\begin{figure}[ht]
    \centering
    \includegraphics[width=\linewidth]{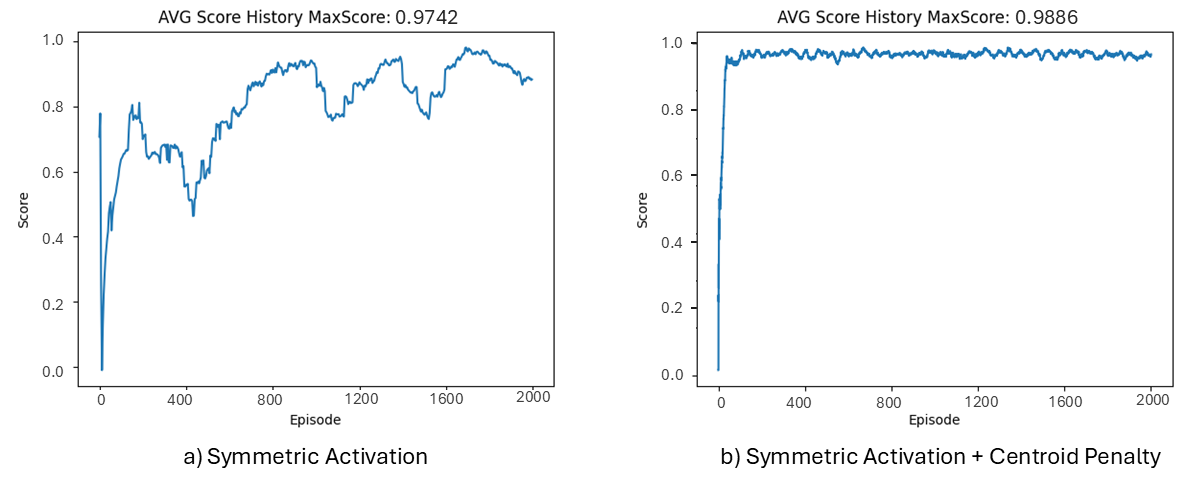}
    \caption{
Evolution of the agent’s performance during training, measured as the moving average of the episode score over the last 100 episodes. (a) Model trained with the symmetric row reward. (b) Model trained with the symmetric reward and additional centroid-based penalty. The smoothed curves highlight overall learning trends and differences in stability and convergence.
}
    \label{fig:score_evolution}
\end{figure}

Figure~\ref{fig:score_evolution} shows the evolution of the agent's performance throughout training, measured as the moving average of the episode score over the last 100 episodes. This smoothed metric allows us to observe general trends in learning while minimizing the effect of short-term fluctuations.

In the case of the first reward function, based solely on symmetric row activation, the agent achieves moderately good performance but fails to consistently reach the maximum possible score. One possible explanation is that the policy overfits to certain permeability configurations and does not generalize well across different randomly generated layouts. The learning curve also exhibits higher variance, with more erratic fluctuations over time, suggesting unstable learning dynamics or sensitivity to episode variability.

By contrast, the second reward function, which adds a penalty based on the vertical displacement of the centroid, produces a more stable and steadily increasing learning curve. The agent is able to generalize more effectively across episodes and approaches the upper bound of the normalized score range. This indicates that incorporating a global symmetry metric, via the centroid, provides a stronger learning signal and improves both convergence and policy robustness.

\subsection{Qualitative Behavior}

\begin{figure}[ht]
    \centering
    \includegraphics[width=\linewidth]{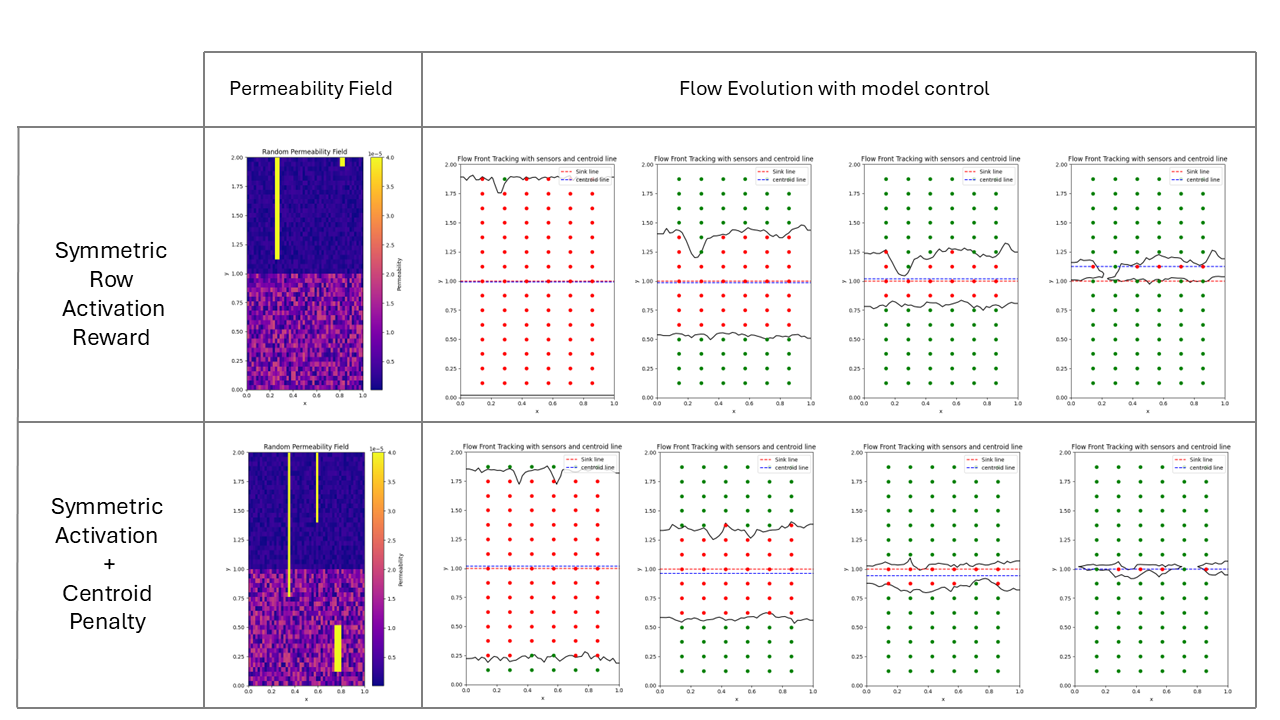}
\caption{
Representative rollouts from both models, each shown with its corresponding permeability field followed by four selected timesteps within the same episode. The top row corresponds to the model trained with the symmetric row reward only, while the bottom row shows the model trained with the symmetric reward and centroid-based penalty. Each timestep snapshot includes the resin front, activated sensors and centroid line, illustrating differences in coordination strategies and convergence behavior.
}
    \label{fig:qualitative_comparison}
\end{figure}

To better understand the behavioral differences induced by the reward function design, Figure~\ref{fig:qualitative_comparison} shows representative rollouts from both models at four selected timesteps throughout an episode. Each row corresponds to a snapshot of the resin front as it progresses through the domain.

In the case of the first reward function, with symmetric row activation only, the agent consistently attempts to activate the domain in a symmetrically balanced manner, progressing row by row from both inlets. However, due to the higher permeability in the lower half of the domain, the bottom flow advances faster. As a result, the agent tends to overshoot near the end of the episode, failing to achieve precise synchronisation at the center.

In contrast, the model trained with the second reward function, which includes a centroid-based penalty, exhibits a more adaptive strategy. While it still follows a roughly symmetric progression during the early and mid stages of the episode, it also compensates for the speed difference by slowing down the faster flow toward the end. This results in a much more synchronised convergence at the center, aligning with the intended objective.

These qualitative differences highlight the effectiveness of incorporating a global balance metric into the reward function, guiding the agent toward more coordinated and goal-aware behaviour.

\section{Limitations and Future perspectives}

While the proposed approach demonstrates promising results in simulation, several limitations must be acknowledged. First, the environment assumes a simplified 2D flow model with binary sensor readings and idealized permeability fields. This model formulation ignores physical phenomena such as variable viscosity, temperature effects, or other stochastic disturbances, which may significantly influence the flow dynamics in real-world applications.

Second, the action space is limited to three discrete control choices and does not allow for a fine-grained modulation of the gate opening, which would be necessary in continuous control scenarios. Additionally, the reward functions are manually engineered and depend on structural symmetry, which may not generalize to domains with irregular geometries or asymmetric boundary conditions.

Finally, although the agent is trained on randomized permeability fields and racetrack configurations, its generalisation capabilities have not been evaluated on distributions significantly different from those seen during training. Robustness under domain shift and deployment in real physical systems remain open challenges.

Several directions remain open to extend and strengthen this work. One natural extension is to scale the system to include more than two inlet gates and more complex domain geometries, which would require multi-agent coordination or higher-dimensional action spaces. Additionally, incorporating noise into the sensor observations or simulating sensor failure scenarios would allow testing the robustness of the learned policy under uncertainty.

From an environment perspective, future work will focus on transitioning toward more realistic simulation models with higher spatial resolution and physical fidelity. This includes using continuous resin flow models, more accurate pressure-velocity coupling, and material properties calibrated from experimental data. Such improvements would allow testing whether the proposed RL framework generalises to realistic industrial settings.

Finally, deploying the trained policies in a physical or hardware-in-the-loop testbed is a long-term goal, which would require addressing challenges related to real-time inference, safety constraints, and domain transfer.

\section{Conclusion}
This work presents an RL approach for synchronised flow control in a two-gate system with heterogeneous and randomised permeability fields. The agent operates in a partially observed environment, relying solely on binary sensor inputs to make discrete control decisions over the inlet gates. Two reward formulations were evaluated: one based on symmetric sensor activation, and another augmented with a centroid-based penalty to guide global flow alignment.

Simulation results demonstrate that the inclusion of a geometric penalty significantly improves synchronisation performance, leading to more stable learning and better convergence across varying spatial layouts. Both quantitative metrics and qualitative rollouts confirm that the second reward formulation enables the agent to adapt its strategy to compensate for flow asymmetries.

The proposed framework highlights the potential of RL to handle spatially complex and dynamic control tasks without requiring explicit physical modelling. These results pave the way for more advanced formulations that incorporate realistic simulation dynamics, richer action spaces, and eventual deployment in physical systems.

\section*{Acknowledgements}
This work has received funding from Horizon Europe, the European Union’s Framework Programme for Research and Innovation, under Grant Agreement No. 101058054 (TURBO). Generalitat Valenciana partially funded this work through project CIPROM/2022/20. 

\bibliographystyle{IEEEtran}
\bibliography{RISO}

\end{document}